\documentclass[letterpaper, 10pt, conference]{IEEEtran}\IEEEoverridecommandlockouts  

\IEEEoverridecommandlockouts
\usepackage{amsmath,amssymb}
\usepackage{graphicx}
\usepackage{booktabs}
\usepackage{xcolor}
\usepackage{cite}
\usepackage[caption=false,font=footnotesize]{subfig}
\usepackage{tikz}
\usetikzlibrary{calc}
\usepackage{hyperref}
\usepackage{enumitem}

\newtheorem{definition}{Definition}
\newtheorem{lemma}{Lemma}

\newcommand{\R}{\mathbb{R}}
\newcommand{\q}{q}
\newcommand{\dq}{\dot q}
\newcommand{\ddq}{\ddot q}

\newcommand{\tb}{p}          \newcommand{\tmax}{\tau_{\max}}
\newcommand{\M}{\mathcal{M}}          \newcommand{\m}{\mathfrak{m}}          \newcommand{\zphi}{\partial_\varphi z_\m}  \newcommand{\zE}{\partial_E z_\m}          

\title{\LARGE \bf
Natural Swing-Up of an Underpowered Double Pendulum
}
\author{Arne Sachtler$^{2,1}$, Manuel Keppler$^{1,3}$, Florian Loeffl$^{1}$, and Alin Albu-Sch\"affer$^{1,2}$\thanks{$^{1}$German Aerospace Center (DLR), Institute of Robotics and Mechatronics, Oberpfaffenhofen, Germany.}\thanks{$^{2}$Technical University of Munich (TUM), Munich, Germany.}\thanks{$^{3}$Eindhoven University of Technology (TU/e), Eindhoven, The Netherlands.}\thanks{Corresponding author: Arne Sachtler \href{mailto:arne.sachtler@tum.de}{\texttt{<arne.sachtler@tum.de>}}.}}

\begin{document}
\bstctlcite{IEEEexample:BSTcontrol}

\maketitle
\begin{abstract}
Robots with weak motors must ride their natural dynamics rather than fight them. We swing up a direct-drive planar two-link robot, whose motors deliver only about 6\,\% of its shoulder gravity torque, along its nonlinear normal modes, two-dimensional invariant manifolds consisting of periodic orbits. On a non-strict mode, injecting energy necessarily moves the state off the manifold, so pumping and stabilization compete for the same bounded torque. We split the input torque into a power-neutral steering component and a scalar pumping input, the only component that changes the energy. The allocation then becomes a linear program in two variables: maximize pumping subject to the torque limits and one control-Lyapunov constraint. The constraint uses the model-based rate of the error, so no stabilizing feedback gain is needed. Experiments in hardware show swing-ups along both modes and compare the allocation with summing-and-clipping and with prioritized stabilization.
\end{abstract}

\section{Introduction}
\label{sec:intro}

We want robots whose wanted motions are natural motions of their dynamics, and nonlinear normal modes (NNMs) are one way to embed them: control uses what is already there instead of forcing new dynamics onto the system. Our control approach holds for $n$ joints and their $n$ modes; we show it on two, whose modes are easy to understand and to visualize.

\begin{figure}
  \centering
  \subfloat[Hardware\label{fig:closeup}]{\includegraphics[height=0.5\columnwidth]{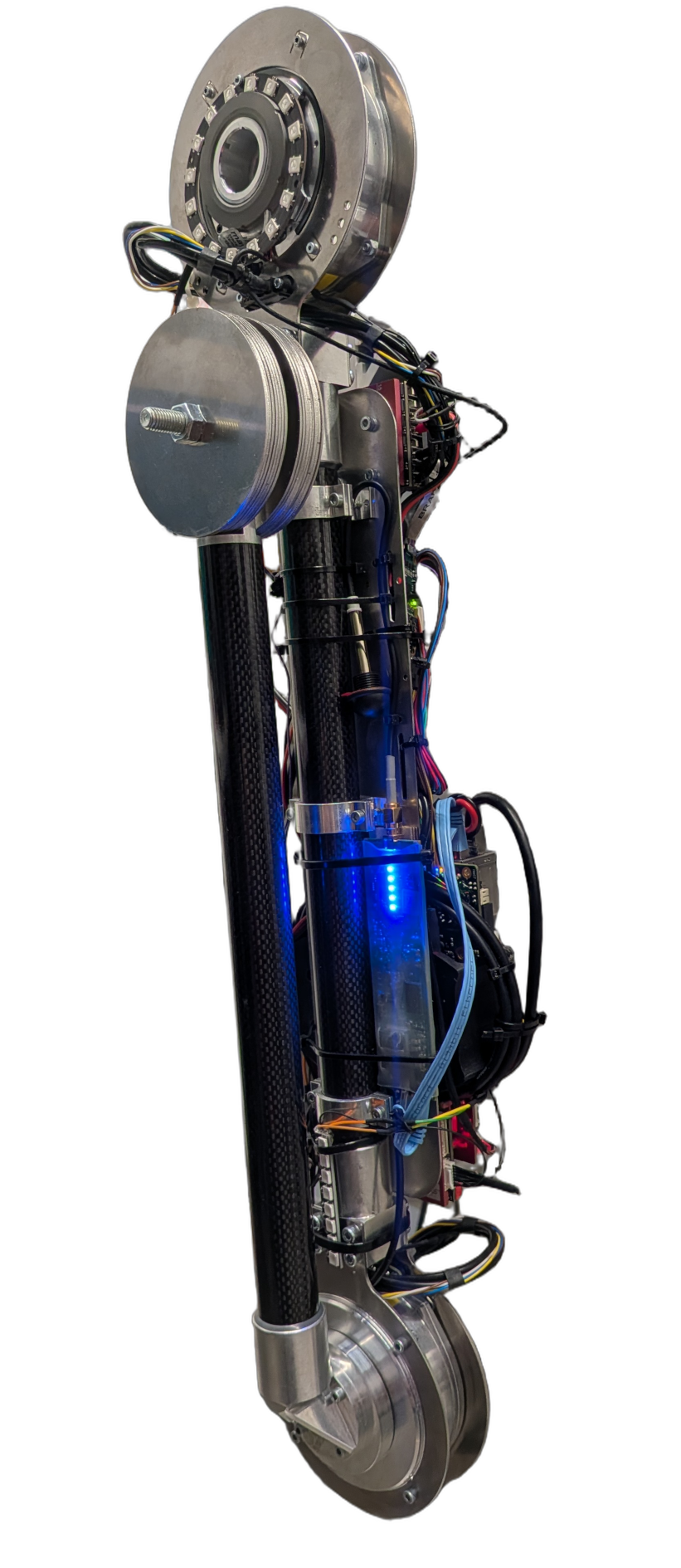}}\hfil
  \subfloat[Full torque\label{fig:fulltorque}]{\includegraphics[height=0.5\columnwidth]{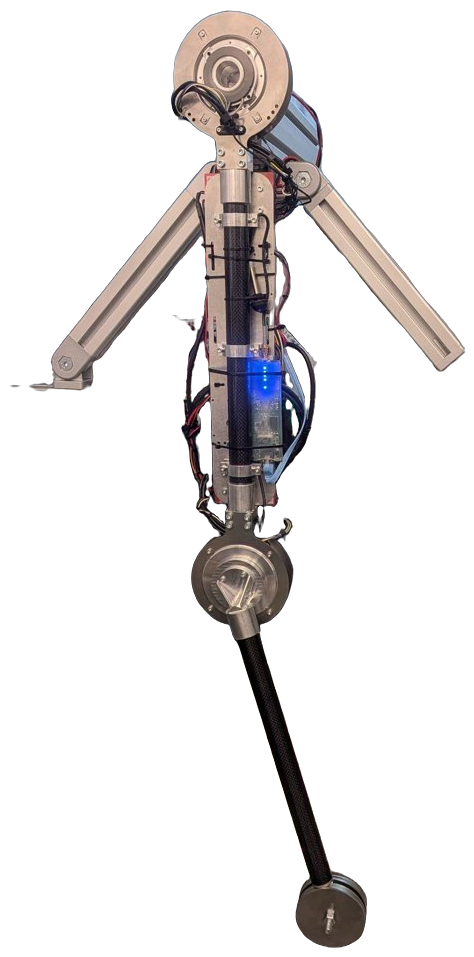}}\hfil
  \subfloat[Model\label{fig:dp}]{\includegraphics[height=0.5\columnwidth]{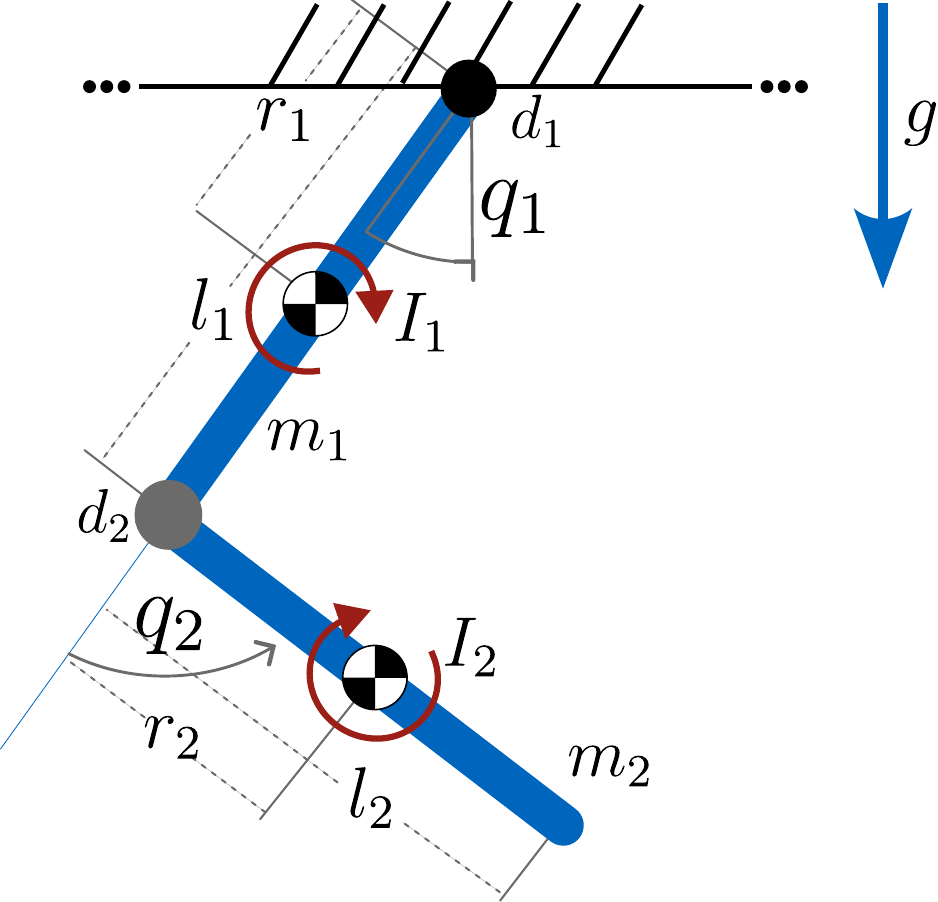}}
  \caption{(a)~Close-up of the robot, link~1 hanging and link~2 upright: direct-drive motors in both joints, electronics and battery on the links. (b)~With full torque in both joints, the model's static pose is $q = (0^\circ, 16.3^\circ)$: link~1 hangs straight down and link~2 tilts by $16.3^\circ$. (c)~Model: link lengths $l_i$, centers of mass $r_i$, masses $m_i$, inertias $I_i$, viscous damping $d_i$; $q_1$ from the downward vertical, $q_2$ relative to link~1.}
  \label{fig:robot}
\end{figure}

Swinging up a robot whose motors cannot even hold its arm horizontal is a matter of timing, not of force. The direct-drive motors of the planar two-link robot of Fig.~\ref{fig:robot} deliver only $1.1$\,Nm, about 6\,\% of the gravity torque in the first joint with the arm held horizontally. With full torque in both joints, link~2 tilts by $16.3^\circ$ and link~1 stays hanging, because the reaction of the elbow torque cancels the shoulder torque (Fig.~\ref{fig:fulltorque}). Neither a joint PD controller nor a controller that tracks a direct path to the upright pose (Fig.~\ref{fig:setup}) can get there; the robot has to collect the energy over many swings. A single pendulum is swung up by energy control, which puts it on the homoclinic orbit that ends upright~\cite{Aastroem2000}. The double pendulum is chaotic~\cite{Shinbrot1992}, so energy control alone most likely ends in chaotic motion, and energy-based swing-ups of double pendulums pass through intermediate equilibria~\cite{Yamakita1995,Aastroem1999}. We instead guide the swing-up along the NNMs of the robot~\cite{Shaw1993,Kerschen2009,AlbuSchaeffer2020}, families of periodic orbits of the free dynamics, one per energy. On one mode the robot behaves much like a single pendulum: there is no chaos, and the modes connect the hanging equilibrium to a homoclinic orbit at the upright one~\cite{Sachtler2024}. Pumping energy along a mode thus carries the robot from rest to almost upright, where a short precomputed trajectory catches it in a handstand.

Like earlier work, we ride a mode with two control loops, one that keeps the state on the mode and one that changes the energy. Both share the same limited motor torques, a box of admissible torques with one bound per joint, which we call the box constraint. Prior work sums the two loops~\cite{Santina2021,Bjelonic2022}, gives priority to stabilization~\cite{Sachtler2024}, or solves a nonlinear MPC in simulation~\cite{Coelho2022}. The loops interfere: a torque that corrects the error with respect to the mode generally does work and changes the energy. On a non-strict mode, one whose orbits do not all run along the same path in configuration space, no bounded torque can change the energy while keeping the state on the mode~\cite[Thm.~1]{Santina2021a}, and the modes of our robot are non-strict (Sec.~\ref{sec:strictness}). Power-neutral steering is known from gyroscopic forces~\cite{Zhang2004}, power-continuous curve tracking~\cite{Duindam2004} and geometric fabrics~\cite{VanWyk2022}, and pumping along the momentum stays on a strict mode~\cite{AlbuSchaeffer2021}. What is not known is how to allocate the two under a box constraint on a non-strict mode, where pumping leaves the mode and steering must undo it.

We split the input torque by its power into a power-neutral steering component and a scalar pumping input, the only component that changes the energy. The allocation then becomes a linear program in two variables: pump as hard as the box allows, subject to one control-Lyapunov function (CLF) constraint~\cite{Ames2014} that makes a transverse error measure decay. Unlike classical control allocation~\cite{Johansen2013}, there is no actuator redundancy; we allocate the torque box between two tasks. No feedback gains are designed; the transverse instability of the orbits sets the decay rate. The PD stabilizers of earlier mode-based swing-ups~\cite{Santina2021,Bjelonic2022,Sachtler2024} and transverse feedback linearization of orbits~\cite{Banaszuk1994,Nielsen2008} need gains tuned against the pumping, and optimal orbit stabilization~\cite{Beck2023} handles a single orbit with unbounded inputs, not a family under a tight box constraint.

\subsection{Contributions}
\begin{itemize}
  \item A gain-free controller that excites and stabilizes a mode with limited torque: one linear program allocates pumping and power-neutral steering under the box constraint. 
  \item Hardware experiments of the swing-up along nonlinear normal modes of~\cite{Sachtler2024}, with the new controller and the identified model of the robot.
  \item A strictness analysis of the modes in the kinetic-energy metric.
\end{itemize}

\section{System and Nonlinear Normal Modes}
\label{sec:prelim}

\subsection{System dynamics}
\label{sec:model}
The robot is a planar double pendulum with direct-drive joints (Fig.~\ref{fig:dp}), with configuration $q$ on the torus $Q = \mathbb{T}^2$, joint velocities $\dq \in T_qQ$ and state $z = (q, \dq) \in TQ$. Table~\ref{tab:params} in Sec.~\ref{sec:platform} gives the identified parameters. This model differs from the idealized one of~\cite{Sachtler2024}, which had, for example, no link inertias about the centers of mass; it matches the hardware, and all modes in this paper are computed for it. The dynamics are
\begin{equation}
  M(q)\ddq + c(q,\dq) + \mathrm{d}V(q) + D\dq = \tau,
  \qquad |\tau_i| \le \tau_{\max,i},
  \label{eq:dyn}
\end{equation}
with the actuator torque $\tau$, the inertia matrix $M$, Coriolis and centrifugal terms $c$, the differential $\mathrm{d}V$ of the gravitational potential $V$, and viscous damping $D = \operatorname{diag}(d_1, d_2)$; the torque bound holds per component. The drift of the free system is
\begin{equation}
  f(z) := \begin{bmatrix} \dq \\ M(q)^{-1}\bigl(-c(q,\dq) - \mathrm{d}V(q)\bigr) \end{bmatrix},
  \label{eq:f}
  \nonumber
\end{equation}
and the energy and its rate are
\begin{equation}
  E = \tfrac12 \dq^\top M(q)\dq + V(q), \qquad
  \dot E = \dq^\top(\tau - D\dq).
  \label{eq:energy}
  \nonumber
\end{equation}
A friction feedforward $\tau_{\mathrm{ff}} = D\dq$ cancels the damping, so the power of the remaining torque is the only thing that changes the energy. Since the friction is always cancelled, we compute the modes for the conservative system.

\subsection{Nonlinear normal modes}
\label{sec:modes}
\begin{figure}[t]
  \centering
  \subfloat[Orbits on the configuration torus\label{fig:torusa}]{\includegraphics[width=\columnwidth]{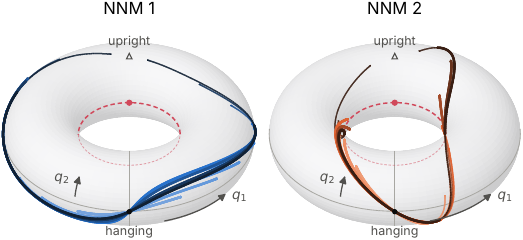}}\\
  \subfloat[Orbits in the kinetic-energy-metric-warped torus\label{fig:torusb}]{\includegraphics[width=\columnwidth]{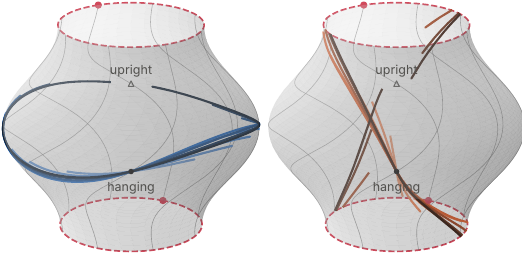}}\\
  \subfloat[Period and transverse instability\label{fig:floquet}]{\includegraphics[width=\columnwidth]{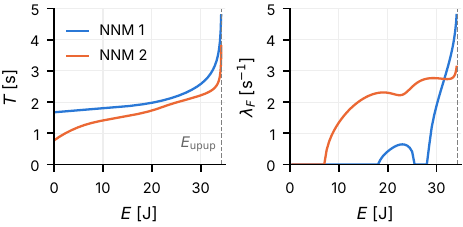}}
  \caption{(a)~Brake orbits of NNM~1 and NNM~2 at 2 to 34\,J (light to dark) on the configuration torus $Q = S^1 \times S^1$, drawn as a donut (NNM~1 left, NNM~2 right, also in (b)); $q_1$ runs around the large circle, $q_2$ around the tube, and $q_2 = 0$ is the outer equator (grey), with the hanging point at the front; arrows: increasing $q_1$, $q_2$; fine dashes: parts behind the surface. (b)~The same torus in the kinetic-energy metric, embedded isometrically as a surface of revolution, one copy $q_2 \in [-\pi,\pi]$; grey: lines $q_1 = \mathrm{const}$. Dashed red in (a) and (b): the cut $q_2 = \pm\pi$, red dot: $(q_1,q_2) = (\pi,\pm\pi)$; in (b) the two rims are the same circle, identified after a rotation by $54.7^\circ$. (c)~Period $T$ and Floquet exponent $\lambda_F$ of the free orbits over their energy; dashed: $E_{\mathrm{upup}}$.}
  \label{fig:torus}
\end{figure}

We use the term nonlinear normal mode in the sense of~\cite{AlbuSchaeffer2020}. An NNM $\M \subset TQ$ is the global version of a Lyapunov subcenter manifold~\cite{Kelley1969,Haller2016}, usually a local construct near an equilibrium. It is a two-dimensional invariant manifold of the free system that collects one continuous family of periodic orbits: $\M$ is foliated by these orbits, one per energy level, and is tangent to a linear mode at the stable equilibrium. By time reversibility the orbits are brake orbits, which come to rest at two turning points~\cite{Wotte2025LSM}. We compute one NNM for each linear mode by continuation~\cite{Sachtler2024}. Both modes are computed up to $E_{\mathrm{upup}} - \delta$, just below the energy $E_{\mathrm{upup}}$ of the upright equilibrium, where the period grows without bound (Fig.~\ref{fig:floquet}).

Fig.~\ref{fig:torusa} shows both modes on the configuration torus. Along NNM~1 the links swing together and $q_2$ stays small, so the orbits run along the outer equator and reach almost all the way around in $q_1$. Along NNM~2 the links swing against each other, and at high energy the elbow turns almost a full revolution.

\begin{definition}[Strict mode]
  \label{def:strict}
  Let $\pi\colon TQ \to Q$ project a state to its configuration. A mode $\M$ is strict if its projection $\pi(\M)$ is a single curve $N \subset Q$.
\end{definition}
Then $\M \subset TN$: all orbits run back and forth along $N$, only further out, and the pumping input $M\dq$ accelerates along $T_qN$ and never leaves $\M$~\cite{AlbuSchaeffer2021}. On a non-strict mode, orbits of different energies take different paths, and pumping leaves $\M$.

\subsection{Energy-phase chart}
\label{sec:chart}
The energy and a phase along the orbit parametrize a mode like polar coordinates: the energy is the radius, the phase the azimuth, and together they give the configuration on the corresponding periodic orbit of the mode; \cite[Ch. 11]{ManifoldBook} visualizes this parametrization. We use the energy $E$ and the phase $\varphi$ as coordinates of a chart $\R \times S^1 \to \M$, $(E,\varphi) \mapsto z_\m = (\eta, \dot\eta)(E,\varphi)$, and store only its configuration part, $q_\m = \eta(E,\varphi)$ (the subscript $\m$ marks a point on $\M$), on a regular $(E,\varphi)$ grid with smooth interpolation. Along an orbit the velocity is parallel to $\partial_\varphi\eta$, and its magnitude follows from the energy, so, with the kinetic-energy norm $\|x\|_M = (x^\top M(q)\,x)^{1/2}$,
\begin{equation}
  \begin{aligned}
  q_\m &= \eta(E,\varphi),\\
  \dq_\m &= \dot\eta(E,\varphi) := \sqrt{2\bigl(E - V(q_\m)\bigr)}\;\frac{\partial_\varphi \eta}{\|\partial_\varphi \eta\|_{M(q_\m)}} .
  \end{aligned}
  \label{eq:query}
  \nonumber
\end{equation}
Two tangent vectors of $\M$ at $z_\m = (\eta, \dot\eta)$ are used throughout,
\begin{equation}
  \zphi = \partial_\varphi(\eta, \dot\eta), \qquad \zE = \partial_E(\eta, \dot\eta),
  \label{eq:tangents}
  \nonumber
\end{equation}
along the orbit and to the neighboring orbit at the same phase. On $\M$ the free flow is tangent to the orbit, $f(z_\m) = \dot\varphi\,\zphi$. As the phase we use $\varphi = \operatorname{atan2}(q_1, \dq_1)$, which increases monotonically along every orbit of both modes of this robot. This is an empirical property, explained in~\cite{Sachtler2024}; where it fails, an arc-length phase with a closest-point lookup serves the same purpose at higher computational cost.

\subsection{Numerical test for strictness}
\label{sec:strictness}
With $a = \partial_\varphi\eta$ and $b = \partial_E\eta$, the chart shows strictness.
\begin{lemma}[Chart test for strictness]
  \label{lem:strict}
  Let $\eta \in C^1$ be a chart of $\M$ for $E > 0$. Then
  \begin{equation}
    \M \text{ strict} \iff \partial_E\eta \in \operatorname{span}\{\partial_\varphi\eta\} \quad \forall (E,\varphi) : \partial_\varphi\eta \neq 0.
    \label{eq:lemma}
  \nonumber
\end{equation}
\end{lemma}
In words, the step $\partial_E\eta$ to the neighboring orbit at the same phase stays tangent to the path of the orbit, and a zero step is trivially tangent. $\partial_E\eta$ vanishes at $q = 0$, which every orbit passes (see below); at the turning points $\partial_\varphi\eta = 0$, and the condition follows there by continuity. A mode is non-strict where $\partial_E\eta$ has a component transverse to $\partial_\varphi\eta$.

Parallelism does not depend on a metric, but the size of the deviation does. The relevant metric is the kinetic-energy metric
\begin{equation}
  \mathrm{d}s^2 = \mathrm{d}q^\top M(q)\,\mathrm{d}q,
  \label{eq:kemetric}
\end{equation}
which assigns a small configuration change $\mathrm{d}q$ the length $\mathrm{d}s$. The test compares tangent vectors, and the kinetic-energy metric is the natural inner product of a mechanical system on them; the Euclidean one in $(q_1, q_2)$ changes with the joint coordinates, e.g.\ from relative to absolute angles. It also measures effort: an acceleration $\alpha$ needs the torque $\tau = M\alpha$, and $\|\alpha\|_M^2 = \tau^\top M^{-1}\tau$, so turning the velocity by an $M$-angle $\theta$ at constant speed takes a torque impulse of size $\|\dq\|_M\,\theta$ in this norm. Locally we measure the $M$-sine of the angle between $a$ and $b$,
\begin{equation}
  \begin{aligned}
  s(E,\varphi) &= \sqrt{1 - \frac{(a^\top M b)^2}{(a^\top M a)(b^\top M b)}} = |\sin\angle_M(a,b)|,\\
  a &= \partial_\varphi\eta,\quad b = \partial_E\eta,
  \end{aligned}
  \label{eq:sloc}
\end{equation}
which vanishes identically iff the mode is strict. Globally we take the highest-energy computed orbit, the top orbit, as reference curve, with kinetic-energy length $L$, and measure the largest kinetic-energy distance $h$ of any point of any lower orbit to it. In a strict mode every lower orbit is a piece of the top orbit's path and $h = 0$; otherwise $h$ is the half-width of the band the orbits form. Both modes of this robot are of Rosenberg type~\cite{Rosenberg1966}: along every orbit both joints come to rest at the same instants and pass through zero at the same instant, so every orbit crosses the hanging configuration $q = 0$. The comparison of the orbit tangents at $q = 0$ in Table~\ref{tab:strict} relies on this; for orbits that miss $q = 0$ it is not defined.

To see these distances, Fig.~\ref{fig:torusb} shows the configuration torus embedded isometrically in $\R^3$ with the metric~\eqref{eq:kemetric}; since $M$ depends only on $q_2$, it is a surface of revolution.

Table~\ref{tab:strict} and Fig.~\ref{fig:torusb} show the result. Both modes are non-strict. In Euclidean $(q_1, q_2)$ coordinates NNM~2 looks stricter, because its orbits rotate by at most $5^\circ$; in the kinetic-energy metric NNM~1 is four times stricter ($h/L = 0.9$\,\% vs.\ $3.6$\,\%). This corrects the impression of~\cite{Sachtler2024}, where the second mode looked stricter in the Euclidean $(q_1, q_2)$ plane.

\begin{table}[t]
  \centering
  \caption{Non-strictness of both modes. The ranking flips between Euclidean $q$ and the kinetic-energy (KE) metric.}
  \label{tab:strict}
  \begin{tabular}{lcc}
    \toprule
    measure & NNM 1 & NNM 2 \\
    \midrule
    orbit tangent at $q=0$ vs.\ linear mode, Euclidean & $45^\circ$ & $5^\circ$ \\
    same, KE metric $M(0)$ & $14.7^\circ$ & $23.0^\circ$ \\
    $s$ \eqref{eq:sloc}, mean over $\varphi$ and $E$ & 0.11 & 0.16 \\
    $s$, peak over $E$ of the mean over $\varphi$ & 0.26 & 0.29 \\
    $h/L$, Euclidean $q$ & 3.6\,\% & 2.7\,\% \\
    $h/L$, KE metric & \textbf{0.9\,\%} & \textbf{3.6\,\%} \\
    \bottomrule
  \end{tabular}
\end{table}

\subsection{Transverse stability}
\label{sec:floquet}
Non-strictness says that pumping pushes the state off $\M$; the stability of the free orbits says whether a state off $\M$ drifts further away. For each periodic orbit on $\M$, with period $T(E)$, we compute the monodromy matrix $\Phi(T)$, which maps a small deviation over one period, by integration along the orbit. It has two unit eigenvalues, along the orbit and toward the neighboring orbit, and a pair $\mu$, $1/\mu$ with $|\mu| \ge 1$ transverse to $\M$. A transverse deviation thus grows by the factor $|\mu|$ per period, on average like $e^{\lambda_F t}$ with the Floquet exponent $\lambda_F(E) = \ln|\mu(E)|/T(E)$. Figure~\ref{fig:floquet} shows $\lambda_F$ for both modes: a controller on NNM~2 stabilizes an unstable orbit from about 7\,J on, on NNM~1 only near the top; $\lambda_F$ peaks at $4.8$\,s$^{-1}$ (NNM~1) and $3.1$\,s$^{-1}$ (NNM~2) at 34.1\,J.

\section{Power-Based Input Split and CLF Allocation}
\label{sec:split}
A swing-up along a mode poses two problems that we want to decouple. Mode stabilization keeps the state on the mode, wherever in the family of orbits the robot currently is. Pumping injects energy, so that the robot climbs from orbit to orbit, and stops at the energy goal. On a non-strict mode the two interfere, and both must share the same bounded torque.

\subsection{Mode stabilization: Lyapunov function and desired decay}
\label{sec:desired}
\label{sec:target}
Every control cycle assigns chart coordinates to the current state $z = (q, \dq)$ with an energy estimator $\hat E(z)$ and a phase estimator $\hat\varphi(z)$. Here we assume both are available; Sec.~\ref{sec:estimation} describes the ones on the robot. The target is the point of $\M$ with these coordinates,
\begin{equation}
  z_\m = (\eta, \dot\eta)\bigl(\hat E(z), \hat\varphi(z)\bigr).
  \label{eq:target}
  \nonumber
\end{equation}
It lies on the current energy level, so the error to it says where the state sits relative to the orbit of its own energy. The error is
\begin{equation}
  e = z - z_\m = \begin{bmatrix} e_q \\ e_{\dq} \end{bmatrix},
  \label{eq:error}
  \nonumber
\end{equation}
where the configuration difference is taken on the torus (Sec.~\ref{sec:estimation}). As a Lyapunov function we use the quadratic form
\begin{equation}
  W(e) = \tfrac12\, e^\top P(q)\, e,
  \label{eq:W}
\end{equation}
with a design parameter $\sigma > 0$, a weight $w > 0$ and
\begin{equation}
  P(q) = \begin{bmatrix} (1+w)\,\sigma^2 M(q) & \sigma M(q) \\ \sigma M(q) & M(q) \end{bmatrix}.
  \label{eq:P}
  \nonumber
\end{equation}
$P$ is positive definite, since its Schur complement is $\tfrac{w}{1+w}M$. With the sliding variable $s = e_{\dq} + \sigma e_q$,
\begin{equation}
  W = \tfrac12\, \|s\|_M^2 + \tfrac12\,w\sigma^2 \|e_q\|_M^2 ,
  \label{eq:Wsliding}
  \nonumber
\end{equation}
the kinetic energy of the sliding error plus a position term. Unlike a function of $e_q$ alone, $s$ has relative degree one, so $\dot W$ is affine in the input, and $s$ carries the input: the gradient $(Ms)^\top$ of $W$ with respect to $e_{\dq}$ is nonzero at $e_{\dq} = 0$ whenever $e_q \neq 0$. No feedback law is designed on $e$. Mode stabilization must bring the state onto $\M$ and keep it there, so the controller only asks $W$ to decay at a rate $\rho > 0$,
\begin{equation}
  \dot W \le -\rho W .
  \label{eq:clf}
\end{equation}

\subsection{Power-based input split}
For $\dq \neq 0$ we split the controller torque as
\begin{equation}
  \begin{aligned}
  \tau &= \tau_{\mathrm{ff}} + \beta\, t(\dq) + \gamma\, \tb,\\
  t(\dq) &= \frac{J\dq}{\|\dq\|},\quad
  \tb = M(q)\dq,\quad
  J = \begin{bmatrix} 0 & -1 \\ 1 & 0 \end{bmatrix},
  \end{aligned}
  \label{eq:split}
  \nonumber
\end{equation}
with the steering $\beta$ along $t$ and the pumping input $\gamma$ along the momentum $\tb$.
Since $\dq^\top t = 0$,
\begin{equation}
  \dq^\top t = 0
  \;\Rightarrow\;
  \dot E = \gamma\, \dq^\top\tb = \gamma\,\|\dq\|_M^2 .
  \label{eq:power}
\end{equation}
The steering $\beta t$ does no work and cannot change the energy, whatever $\beta$ is. The pumping input $\gamma\tb$ is the only energy-changing part. The two directions span $T^*_qQ \cong \R^2$ whenever $\dq \neq 0$, because $\dq^\top M\dq > 0$, so the split is a state-dependent change of input coordinates.

\begin{figure}[t]
  \centering
  \subfloat[Split\label{fig:split}]{\raisebox{0.4cm}{\begin{tikzpicture}[scale=0.8,>=latex]
    \draw[gray!60] (0,0) circle (1.4);
    \draw[gray!60,->] (-1.8,0) -- (2.0,0) node[right,black] {\small $v_1$};
    \draw[gray!60,->] (0,-1.6) -- (0,1.9) node[above,black] {\small $v_2$};
    \coordinate (V) at (40:1.4);
    \draw[very thick,->] (0,0) -- (V) node[midway,above left] {$v$};
    \draw[thick,->,blue!70!black] (V) -- (40:2.1) node[right] {\small $\gamma$};
    \draw[thick,->,orange!85!black] (V) -- ($(V)+(130:0.75)$) node[above] {\small $\beta$};
  \end{tikzpicture}}}\hfil
  \subfloat[Allocation program\label{fig:uplane}]{\includegraphics[width=0.58\columnwidth]{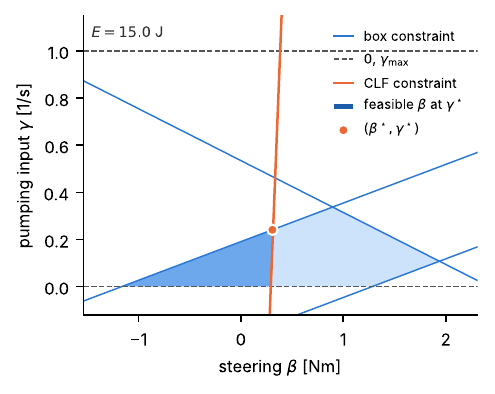}}
  \caption{(a)~The split in kinetic-energy coordinates $v = M^{1/2}\dq$: the pumping input $\gamma$ stretches the velocity, the steering $\beta$ rotates it at constant kinetic energy (circle). (b)~The program \eqref{eq:lp} in the $u = (\beta,\gamma)$ plane at one state of a simulated swing-up along NNM~1 ($E = 15.0$\,J): box constraint of both joints, bounds on $\gamma$ (dashed), CLF constraint (orange), $\mathcal U(z)$ (light), feasible set (dark), optimum and tie interval. The CLF constraint is active: the error, not the box, limits the pumping.}
  \label{fig:alloc}
\end{figure}

In the kinetic-energy metric the pumping input accelerates along the velocity, $M^{-1}\tb = \dq$, while the steering accelerates $M$-orthogonally to it, $\langle\dq, M^{-1}t\rangle_M = 0$. In coordinates $v = M^{1/2}\dq$ the pumping input stretches the velocity vector and the steering rotates it at constant kinetic energy (Fig.~\ref{fig:split}). With the friction cancelled, the robot is control-affine in $u = (\beta, \gamma)$,
\begin{equation}
  \dot z = f(z) + g(z)\,u,
  \;\;
  g(z) = \begin{bmatrix} 0 & 0 \\ M^{-1} t & M^{-1}\tb \end{bmatrix},
  \;\;
  u = \begin{bmatrix} \beta \\ \gamma \end{bmatrix}.
  \label{eq:affine}
\end{equation}

\subsection{Error dynamics and lookahead}
\label{sec:edot}
The rate of $W$ needs $\dot e = \dot z - \dot z_\m$, and the target moves with the state through both chart coordinates. By the chain rule,
\begin{equation}
  \dot z_\m = \zphi\, \dot{\hat\varphi} + \zE\, \dot E,
  \qquad
  \dot{\hat\varphi} = \nabla\hat\varphi(z)^\top \dot z :
  \label{eq:targetvel}
  \nonumber
\end{equation}
the pumping input $\gamma$ moves the target to the next orbit, and because the phase is recomputed from the state, the target slides along the orbit whenever the state moves. Collecting the terms in $\dot z$ gives the projection
\begin{equation}
  \Pi = I - \frac{\zphi\,\nabla\hat\varphi^\top}{\nabla\hat\varphi^\top\zphi}, \qquad \Pi\, \zphi = 0,
  \label{eq:Pi}
  \nonumber
\end{equation}
which removes the part of $\dot z$ that only moves the target along the orbit. The energy is a particularly simple output: $L_f E = 0$ and $L_g E = (0,\, \dq^\top\tb)$, with the Lie derivative $L_f E = \nabla E^\top f$, so it is invariant to $\beta$ and has relative degree one in $\gamma$. $\Pi$ must be built from the phase estimator that is actually used (Sec.~\ref{sec:estimation}). Inserting~\eqref{eq:affine}, the free flow on $\M$ can be subtracted for free,
\begin{equation}
  f(z_\m) = \dot\varphi\,\zphi \;\Rightarrow\; \Pi f(z) = \Pi\bigl(f(z) - f(z_\m)\bigr),
  \label{eq:drift}
  \nonumber
\end{equation}
so the free dynamics enter only as the difference between where the robot is and where it should be. The result is
\begin{equation}
  \dot e = \underbrace{\Pi\bigl(f(z) - f(z_\m)\bigr)}_{F(z)}
  + \underbrace{\bigl(\Pi\, g(z) - \zE\, L_g E(z)\bigr)}_{G(z)}\, u .
  \label{eq:edot}
\end{equation}
These three terms are a model-based lookahead: the allocation reacts to where the error is heading, not only to where it is. The drift difference predicts how gravity and Coriolis forces pull state and target apart, which matters most near the top, where a small offset means a large difference in gravity torque. The term in $\zE$ predicts that pumping drags the target to the next orbit, and $\Pi$ that the target slides along with the state.

With~\eqref{eq:W} and~\eqref{eq:edot}, the rate of $W$ under the split input is affine in $u$,
\begin{equation}
  \begin{aligned}
  \dot W &= e^\top P\,\dot e + \tfrac12\, e^\top \dot P\, e\\
  &= \underbrace{e^\top P\, F(z) + \tfrac12\, e^\top \dot P\, e}_{c_0}
  + \underbrace{e^\top P\, G(z)}_{[c_\beta\;\; c_\gamma]}\, u ,
  \end{aligned}
  \label{eq:Wdot}
  \nonumber
\end{equation}
where $\dot P$ contains $\dot M = \partial_q M[\dq]$, which depends on the state only, so the rate stays affine in $u$.\footnote{Written out, $c_\beta = e^\top P\, \Pi\, (0,\, M^{-1}t)$ and $c_\gamma = e^\top P\, \Pi\, (0,\, M^{-1}\tb) - e^\top P\, \zE\, \dq^\top\tb$, with $(0, x)$ the stacked vector.}
$c_0$ is the rate of $W$ without input, $c_\beta$ the leverage of the steering, and $c_\gamma$ the effect of the pumping input, including the motion of the target. On a non-strict mode $c_\gamma$ is generically nonzero: pumping itself pushes the state off $\M$. For a fixed state the desired dynamics~\eqref{eq:clf} become one half-plane in the $u$-plane, the CLF constraint
\begin{equation}
  c_0 + c_\beta\,\beta + c_\gamma\,\gamma \le -\rho W .
  \label{eq:clfc}
\end{equation}

\subsection{Allocation program}
\label{sec:program}
Two further limits act on $u$. The actuators bound each joint torque, $|\tau_i| \le \tau_{\max,i}$ in~\eqref{eq:dyn}; with the split this box constraint becomes a bound on $\tau_{\mathrm{ff}} + \beta t + \gamma\tb$. The pumping input is capped, $0 \le \gamma \le \gamma_{\max}$: the lower bound keeps it from braking, and the upper bound limits the pumping rate so that pumping stops at $E_{\mathrm{des}}$. Together they form a state-dependent polygon in the $u$-plane,
\begin{equation}
  \mathcal{U}(z) = \bigl\{u : |\tau_{\mathrm{ff}} + [t\;\; \tb]\,u| \le \tmax,\;\; 0 \le \gamma \le \gamma_{\max}\bigr\},
  \label{eq:U}
\end{equation}
bounded because $\|t\| = 1$ and $\gamma$ is capped. With $\gamma \ge 0$ the pumping input never brakes, so by~\eqref{eq:power} the energy never decreases. The box constraint is physics, and monotone energy is what the swing-up is for, so both are hard. The CLF constraint is soft: when no admissible input meets it, the controller comes as close as it can and reports by how much it misses. This order distinguishes the approach from prioritized stabilization~\cite{Sachtler2024}, which makes the stabilizer hard and the energy soft. The allocation maximizes pumping over the polygon cut by the CLF constraint,
\begin{align}
  \gamma^\star = \max_{\beta,\gamma}\;\; & \gamma \label{eq:lp}\\
  \text{s.t.}\;\; & |\tau_{\mathrm{ff}} + \beta\,t + \gamma\,\tb| \le \tmax \nonumber\\
  & 0 \le \gamma \le \gamma_{\max} \nonumber\\
  & c_0 + c_\beta\,\beta + c_\gamma\,\gamma \le -\rho W. \nonumber
\end{align}
It is a CLF program in the sense of~\cite{Ames2014}, with maximal pumping as the objective instead of a minimal input norm, and the box-constraint polygon as the input set: an LP in two variables over seven half-planes, solved in closed form by vertex enumeration. The steering is the free variable that lets $\gamma$ go high: if pumping pushes $W$ up ($c_\gamma > 0$), the program picks the steering that compensates, as long as the box allows. At $\gamma^\star$ the steering is often not unique, and we take the least steering that still meets the CLF constraint,
\begin{equation}
  \begin{aligned}
  \beta^\star = \arg\min\bigl\{|\beta| :\ & (\beta,\gamma^\star) \in \mathcal{U}(z),\\
  & c_0 + c_\beta\beta + c_\gamma\gamma^\star \le -\rho W\bigr\},
  \end{aligned}
  \label{eq:tie}
  \nonumber
\end{equation}
so the robot steers only as much as the CLF constraint demands. Choosing the steepest decrease instead is bang-bang: the steering jumps between the box edges. The CLF constraint can miss the polygon entirely: then no admissible input makes $W$ decay at rate $\rho$. In that case the controller takes the vertex of $\mathcal U(z)$ with the smallest predicted $\dot W$ and counts the cycle as infeasible. If pumping makes the error worse, that vertex usually has $\gamma = 0$, so pumping pauses until the error is under control. Fig.~\ref{fig:uplane} shows the program at one state.

\subsection{Choice of the rate}
\label{sec:rho}
The rate $\rho$ is the one tuning parameter of the CLF constraint, and the free dynamics set its lower bound. A transverse error grows like $e^{\lambda_F t}$ with the Floquet exponent of Sec.~\ref{sec:floquet}, so $W$, quadratic in the error, grows like $e^{2\lambda_F t}$. In the cycles where the constraint is infeasible nothing stops this growth, and the feasible cycles must undo it. The demanded rate must therefore outpace the transverse instability of every orbit of the mode that the swing-up passes, so $\rho$ can be decided from the Floquet exponents of the mode's orbits alone:
\begin{equation}
  \rho > 2\bar\lambda_F, \qquad \bar\lambda_F = \max_{E \le E_{\mathrm{des}}} \lambda_F(E).
  \label{eq:rhorule}
\end{equation}
The rule is a design rule, not a guarantee: a faster rate leaves less of the box for pumping and makes the constraint infeasible more often, and quickly passed orbits may need less.

The rate in turn bounds $\sigma$ from below. On $s = 0$ the gradient of the sliding term $\tfrac12\|s\|_M^2$ vanishes, so the input has no leverage, and with $e_{\dq} = -\sigma e_q$ the position term decays on its own, $\dot W = w\sigma^2 e_q^\top M e_{\dq} = -2\sigma W$ (up to the small $\dot M$ term), for every $w$. The constraint is then feasible on $s = 0$ only if
\begin{equation}
  \rho < 2\sigma .
  \label{eq:lamrule}
  \nonumber
\end{equation}
The design thus runs $\lambda_F \to \rho \to \sigma$.

\section{Implementation}
\label{sec:impl}

\subsection{Robot}
\label{sec:platform}

\begin{figure}[t]
  \centering
  \includegraphics[width=0.85\columnwidth]{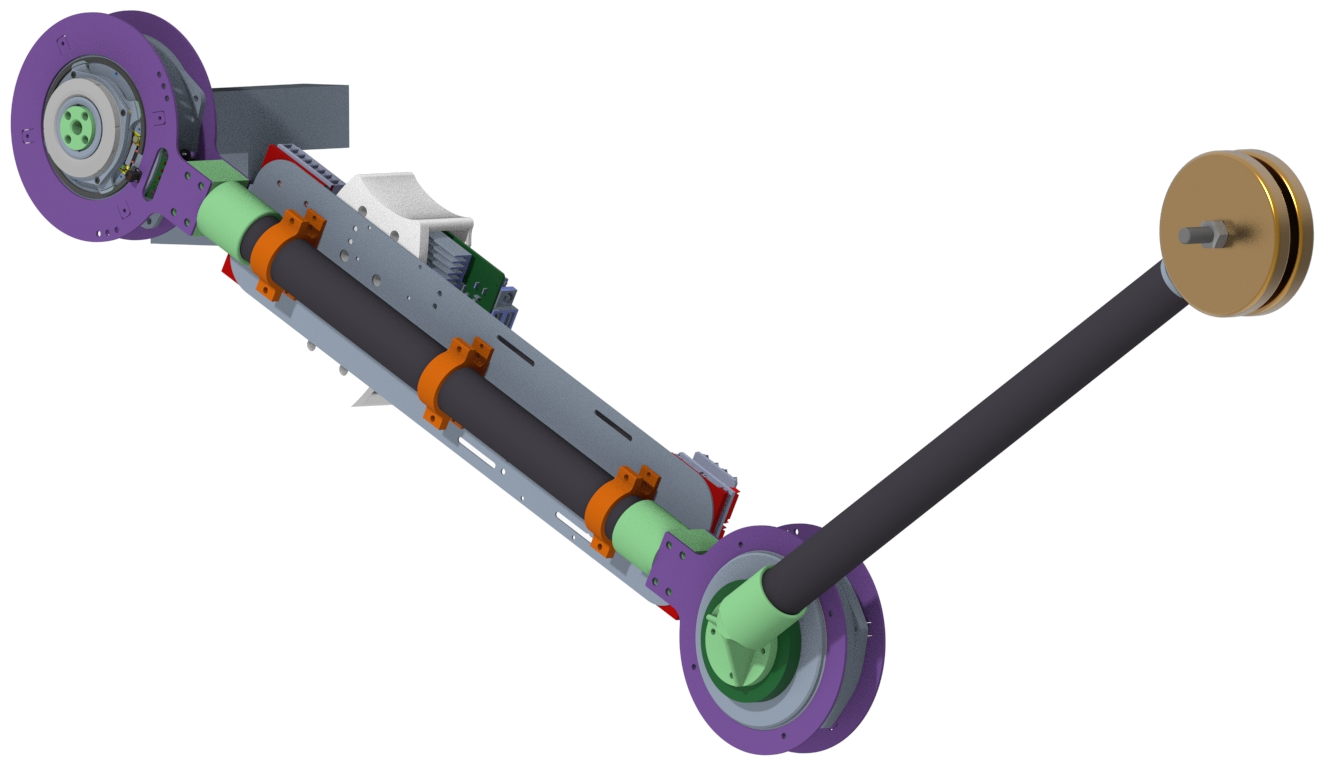}
  \caption{CAD model of the robot: shoulder motor (left), link~1 with the electronics, elbow motor, and link~2.}
  \label{fig:cad}
\end{figure}
\begin{table}[t]
  \centering
  \caption{Robot parameters (identified).}
  \label{tab:params}
  \begin{tabular*}{\columnwidth}{@{}l@{\extracolsep{\fill}}ccl@{}}
    \toprule
    parameter & link 1 & link 2 & unit \\
    \midrule
    length $l_i$ & 0.5 & 0.408 & m \\
    mass $m_i$ & 3.143 & 1.078 & kg \\
    center of mass $r_i$ & 0.257 & 0.370 & m \\
    inertia $I_i$ & 0.135 & 0.008 & kg\,m$^2$ \\
    damping $d_i$ & 0.018 & 0.007 & Nms/rad \\
    torque limit $\tmax$ & 1.1 & 1.1 & Nm \\
    gravity torque, horizontal & 17.12 & 3.91 & Nm \\
    \midrule
    hanging energy $E_{\mathrm{hang}}$ & \multicolumn{2}{c}{0.00} & J \\
    upright energy $E_{\mathrm{upup}}$ & \multicolumn{2}{c}{34.25} & J \\
    \bottomrule
  \end{tabular*}
\end{table}
The robot (Figs.~\ref{fig:closeup} and~\ref{fig:cad}) is truly direct-drive, without gears. Each joint has a TQ ILM 85x13 motor and a Synapticon Node~400 driver. The joint position comes from the motor's magnetic position sensor, which the driver also uses for commutation, with 18\,bit per turn ($2.4\cdot10^{-5}$\,rad). There is no torque sensor; the torque follows from the model-based current-to-torque relation of the motor, which is accurate without a gearbox. An UP\textsuperscript{2} board runs the controller and the EtherCAT link to the drivers at 1\,kHz. A battery on the robot and no wiring across the joints let both joints turn without limit. Every command passes a first-order low-pass at 20\,Hz.

The controller is implemented in C++. It uses CGAL~\cite{cgal} for the meshing and interpolation of the NNMs, links and nodes~\cite{Schmidt2026} as middleware, and Pinocchio~\cite{Carpentier2019} for the dynamics.

\subsection{Estimation and target}
\label{sec:estimation}
An unscented Kálmán filter on the encoder positions with the model of Sec.~\ref{sec:model} estimates the state and gives the energy estimate $\hat E$. The phase estimate $\hat\varphi$ is the atan2 phase of Sec.~\ref{sec:chart}, evaluated on the estimated state. Its gradient, which $\Pi$ needs, is exact and in closed form,
\begin{equation}
  \hat\varphi = \operatorname{atan2}(q_1, \dq_1), \qquad
  \nabla\hat\varphi = \frac{(\dq_1,\, 0,\, -q_1,\, 0)}{q_1^2 + \dq_1^2},
  \label{eq:p02}
  \nonumber
\end{equation}
singular only at rest. The configuration part of the error is taken on the torus, $e_q = q \ominus q_\m$, where $\ominus$ subtracts and wraps each component to $(-\pi, \pi]$.

\subsection{Controller details}
\label{sec:ctrldetails}
At the turning points of a brake orbit, $\dq = 0$, the direction $t$ is undefined and $M\dq = 0$, so neither input has authority. In a band $\|\dq\| < \varepsilon$ the controller applies only $\tau_{\mathrm{ff}}$. The energy goal enters through the cap of the pumping input,
\begin{equation}
  \begin{aligned}
  &\gamma_{\max} = \bar\gamma\, r(\hat E),\\
  &r(E) = \min\bigl\{1,\, \max\{0,\, k_E (E_{\mathrm{des}} - E)\}\bigr\}:
  \end{aligned}
  \label{eq:pumping}
  \nonumber
\end{equation}
$r$ is one far below the goal, falls linearly to zero over the last $1/k_E$ joules, and stays zero above it. Letting $\gamma$ carry the sign instead would make the allocation jump between full pumping and full braking whenever the energy estimate crosses $E_{\mathrm{des}}$. The rate follows from~\eqref{eq:rhorule} and Fig.~\ref{fig:floquet}: $2\bar\lambda_F = 9.6$\,s$^{-1}$ on NNM~1 and 6.3\,s$^{-1}$ on NNM~2, both at 34.1\,J. With a margin of about 20\,\% we use $\rho = 12$\,s$^{-1}$ on NNM~1 and $\rho = 8$\,s$^{-1}$ on NNM~2, and $\sigma = 7$\,s$^{-1}$ and $w = 0.1$ for both modes, which satisfies $\rho < 2\sigma$. The remaining parameters are $\tmax = 1.1$\,Nm per joint, $k_E = 1$\,J$^{-1}$, $\bar\gamma = 10$\,s$^{-1}$ and $\varepsilon = 0.1$\,rad/s.

\section{Experiments}
\label{sec:exp}

\begin{figure*}[t]
  \centering
  \subfloat[Upright\label{fig:setup}]{\includegraphics[height=0.245\textwidth]{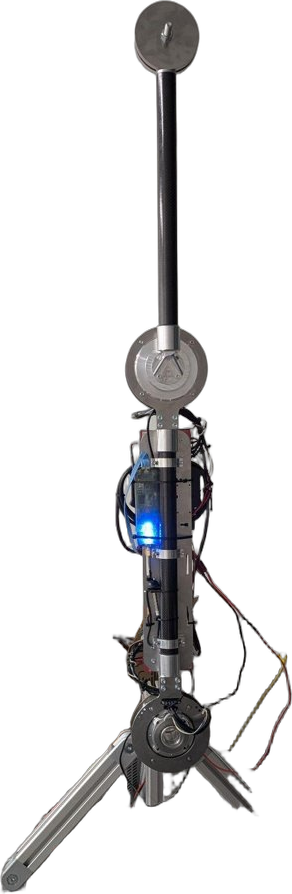}}\hfill
  \subfloat[Tip path, NNM~1\label{fig:tip1}]{\includegraphics[height=0.245\textwidth]{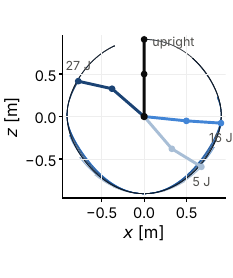}}\hfill
  \subfloat[Tip path, NNM~2\label{fig:tip2}]{\includegraphics[height=0.245\textwidth]{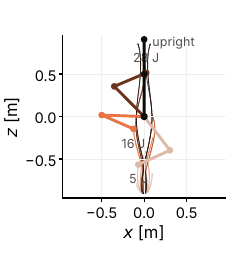}}\hfill
  \subfloat[Energy, Lyapunov function and pumping input\label{fig:modestime}]{\includegraphics[height=0.245\textwidth]{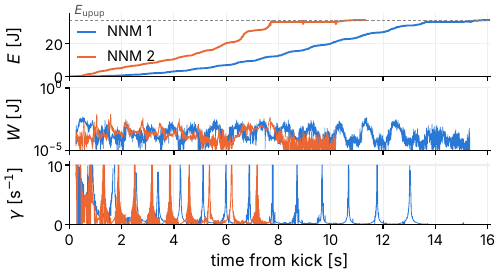}}
  \caption{Swing-up along NNM~1 and NNM~2 with the same controller, each with its rate from \eqref{eq:rhorule}. (a)~The robot upright. (b), (c)~Path of the tip from the kick to the end of the run, shaded by energy, with the robot at turning points and the final upright pose after the catch (black). (d)~Energy $E$, Lyapunov function $W$ and pumping input $\gamma$ of both runs.}
  \label{fig:modes}
\end{figure*}
\subsection{Task and protocol}
\label{sec:protocol}
A run starts at rest. A constant torque for 0.25\,s in the linear modal direction leaves the rest state, where the split is undefined. 
We then employ the controller of Sec.~\ref{sec:program} using the chart $\eta$ of the chosen mode.
The controller then stabilizes the mode and pumps energy to $E_{\mathrm{des}} = 34.1$\,J, which is just below the critical energy $E_{\mathrm{upup}} = 34.25$\,J of the upright pose. The robot then reaches a periodic orbit that almost hits the upright pose.
We then wait for the next turning point, where $\dq = 0$, and switch to a catch controller that stabilizes the upright pose $\q_{d} = (\pi, 0)$ with a simple PD controller.
A failsafe stops the run as soon as the total energy $E$ drops by $1$\,J which our energy-monotone controller should never cause.
All runs use the parameters of Sec.~\ref{sec:ctrldetails}. We first show the swing-up along both modes, then compare our allocation with two baselines.

\label{sec:metrics}For each run we report the time from the kick to 33\,J and the fraction of cycles above 20\,J in which the CLF constraint is infeasible. The comparison with the baselines (Sec.~\ref{sec:baselines}) stops before the catch: every method pumps only to $E_{\mathrm{des}} = 33.1$\,J, just above the 33\,J at which the metrics end. For this comparison we also measure how close each method stays to $\M$. So that all methods are judged alike, the distance is taken to one reference chart of the mode, in the kinetic-energy metric of $W$: with $e_q = q \ominus \eta_{\mathrm{ref}}$ and $e_{\dq} = \dq - \dot\eta_{\mathrm{ref}}$ at $(E(z), \varphi)$, $d(z) = \min_{\varphi} (\sigma^2 e_q^\top M e_q + e_{\dq}^\top M e_{\dq})^{1/2}$. We report its median and 95th percentile for $25 \le E \le 33$\,J.

\subsection{Swing-up along both modes}
\label{sec:hw}
Fig.~\ref{fig:modes} shows one swing-up from rest along each mode; only the chart and $\rho$ differ. NNM~1 reaches 33\,J after 13.6\,s, NNM~2 after 7.7\,s, and in both runs the catch brings the robot to rest upright. The tip paths show the two mode shapes: along NNM~1 the links swing together and the tip sweeps a widening arc, along NNM~2 they swing against each other and the tip stays close to the vertical. The pumping input reaches its cap near the turning points, where its torque $\gamma \tb$ is small; in between, the box or the CLF constraint limits it. On NNM~2 the energy shows small dips, which the model excludes. We attribute them to unmodeled effects of the fast whipping motion of this mode: the bearing forces predicted by the model peak at about 400\,N on NNM~2 against 150\,N on NNM~1, so the friction grows beyond the feedforward, and the suspension of the robot deforms slightly and takes out energy. Above 20\,J the CLF constraint is infeasible in 1.1\,\% of the cycles on NNM~1 and in 5.0\,\% on NNM~2.

\subsection{Baselines}
\label{sec:baselines}
Both baselines use the same target, the same pumping direction and the same ramp, so they differ only in how stabilization and pumping share the box. Both add the PD stabilizer of~\cite{Santina2021,Sachtler2024},
\begin{equation}
  \tau_\m = M(q)\bigl[k_p (q_\m \ominus q) + k_d (\dq_\m - \dq)\bigr],
  \qquad k_d = 2\zeta\sqrt{k_p},
  \label{eq:stab}
  \nonumber
\end{equation}
with $\zeta = 1$ and $k_p$ scheduled on the energy: $k_p = 10$ below 5\,J and 20 above. Summing and clipping~\cite{Santina2021,Bjelonic2022} uses a fixed pumping input $\gamma_c$,
\begin{equation}
  \tau = \mathrm{sat}_{\tmax}\bigl(\tau_{\mathrm{ff}} + \tau_\m + \gamma_c\, r(E)\,\tb\bigr),
  \label{eq:clip}
  \nonumber
\end{equation}
with $\gamma_c = 0.1$ above 5\,J. Clipping changes the direction of the torque, so neither loop gets what it asked for. This baseline is called \emph{sum and clip}. The prioritized scheme~\cite{Sachtler2024} gives the pumping input what the box leaves,
\begin{equation}
  \begin{aligned}
  \tau &= \tau_{\mathrm{ff}} + \tau_\m + \gamma^\star \tb, \\
  \gamma^\star &= \max\{\gamma \in [0,\bar\gamma\, r(E)] : |\tau_{\mathrm{ff}} + \tau_\m + \gamma\tb| \le \tmax\}
  \end{aligned}
  \label{eq:prio}
  \nonumber
\end{equation}
and is called \emph{prio} as it prioritizes the stabilizer over pumping. Prio needs this hand-tuned schedule on NNM~1.

\subsection{Comparison with the baselines}
\label{sec:comparison}
\begin{figure}[t]
  \centering
  \includegraphics[width=\columnwidth]{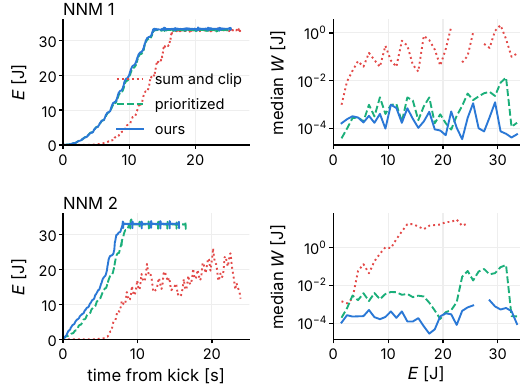}
  \caption{The three controllers from rest to $E = 33$\,J on both modes. Left: energy. Right: median of $W$~\eqref{eq:W} per 1\,J of energy.}
  \label{fig:baselines}
\end{figure}
\begin{table}[t]
  \centering
  \caption{Comparison with the baselines (--: 33\,J not reached)}
  \label{tab:exp}
  \begin{tabular*}{\columnwidth}{@{}l@{\extracolsep{\fill}}ccc@{}}
    \toprule
    method & $t(E{=}33\,\mathrm{J})$ [s] & $d$ [$\sqrt{\mathrm{J}}$] & infeas. \\
    \midrule
    \multicolumn{4}{@{}l}{\emph{NNM~1}} \\
    sum and clip \eqref{eq:clip} & 17.4 & 0.43 / 1.06 & n/a \\
    prioritized \eqref{eq:prio} & 13.7 & 0.08 / 0.19 & n/a \\
    ours & 13.6 & 0.05 / 0.10 & 1.1\,\% \\
    \midrule
    \multicolumn{4}{@{}l}{\emph{NNM~2}} \\
    sum and clip \eqref{eq:clip} & -- & -- & n/a \\
    prioritized \eqref{eq:prio} & 8.8 & 0.39 / 0.57 & n/a \\
    ours & 8.0 & 0.07 / 0.15 & 10.8\,\% \\
    \bottomrule
\end{tabular*}
 \end{table}
Fig.~\ref{fig:baselines} and Table~\ref{tab:exp} compare the three allocations. On NNM~1, the prioritized scheme reaches 33\,J about as fast as ours, 13.7\,s against 13.6\,s, but stays farther from $\M$: between 25 and 33\,J its median distance is 0.08 against 0.05 for ours. It gets there only with a hand-tuned gain schedule (Sec.~\ref{sec:discussion}). Summing and clipping takes 17.4\,s, and its distance grows above 25\,J to a median of 0.43. On NNM~2, ours reaches 33\,J after 8.0\,s and the prioritized scheme after 8.8\,s, with median distances of 0.07 and 0.39; for summing and clipping we found no working parameter set, and its energy peaks at 26\,J.

\subsection{Discussion}
\label{sec:discussion}

The prioritized baseline comes close to our method on NNM~1, but only because its stabilizer gain was scheduled on the energy by hand. We found no single $k_p$ that works for the whole swing-up. With a high $k_p$, the PD stabilizer does work: it takes energy out at low energies as fast as the pumping input puts it in, and the energy levels off far below the goal. With a lower $k_p$ the swing-up makes progress, but the stabilizer no longer holds the state near $\M$ at high energies. Our method has no such gain to tune: the steering is power-neutral and cannot cancel the pumping, $\rho$ follows from the Floquet exponents by~\eqref{eq:rhorule}, and $\sigma$ from $\rho$.

The CLF constraint is not always feasible, and on a non-strict mode it cannot be; the method reports the miss, and the infeasible fraction is a health signal. At the turning points neither input has authority, and no analysis covers the band. The family ends in a homoclinic orbit, so the catch needs a separate controller. The CLF constraint uses the predicted $\dot W$ at the estimated state; the estimated energy is not exactly the energy of the estimated state, and the lag of the torque filter is not modeled. Finally, the allocation is greedy: it maximizes the pumping rate at each instant, which is energy-efficient by construction but not time-optimal over the whole swing-up.

\section{Conclusion}
We split the torque of a weakly actuated robot into workless steering and a scalar pumping input, and allocate both by a two-variable LP under the box constraint and one CLF constraint, whose model-based error rate replaces a stabilizer and its gains. By construction the box holds, the energy never decreases or passes its goal, and the control power is never negative. On hardware, with motors at 6\,\% of the shoulder gravity torque, the controller swings the robot up along both modes, reaching 33\,J after 13.6\,s and 7.7\,s, and stays closer to the mode than both baselines. The method extends directly to more joints, since exactly one torque direction, $M\dq$, changes the energy. Future work includes hardware with more joints, a phase-free stabilizer, and families not rooted in equilibria.

\section*{Acknowledgment}
This work was supported by the European Union through the ERC Proof of Concept grant \emph{SwingBots} (101188218). We thank Harald Wagner, Alexander Kirst, and Martin Görner for help with the hardware. The AI assistant Claude (Anthropic), Model \texttt{Opus 5.5} was used in the writing process.

\bibliographystyle{IEEEtran}

\end{document}